\documentclass[letterpaper, 10 pt, conference]{ieeeconf}  % Comment this line out if you need a4paper

\usepackage[OT1]{fontenc} 

\usepackage{cite}
\usepackage{amsmath,amssymb,amsfonts}
\usepackage{graphicx}
\usepackage{textcomp}
\usepackage{multirow}
\usepackage{multicol}
\usepackage[dvipsnames]{xcolor}
\usepackage{hyperref}
\usepackage{placeins}
\usepackage{soul}
\usepackage{algorithm}
\usepackage{algpseudocode}
\usepackage{amssymb}
\usepackage{bm}

\let\labelindent\relax
\usepackage{enumitem}

\algnewcommand\algorithmicforeach{\textbf{for each}}
\algdef{S}[FOR]{ForEach}[1]{\algorithmicforeach\ #1\ \algorithmicdo}

\title{\LARGE \bf
Expert Composer Policy: Scalable Skill Repertoire for Quadruped Robots
}

\begin{document}

\author{Guilherme Christmann*, Ying-Sheng Luo*, Wei-Chao Chen \\
        Inventec Corporation, Taipei, Taiwan \\ 
        \textit{\{guilherme.christmann, luo.ying-sheng, chen.wei-chao\}@inventec.com}}

\maketitle

\begingroup\renewcommand\thefootnote{*}
\footnotetext{These authors contributed equally, listed alphabetically by last name.}
\endgroup

\thispagestyle{empty}
\pagestyle{empty}

%%%%%%%%%%%%%%%%%%%%%%%%%%%%%%%%%%%%%%%%%%%%%%%%%%%%%%%%%%%%%%%%%%%%%%%%%%%%%%%%

\begin{abstract}
We propose the expert composer policy, a framework to reliably expand the skill repertoire of quadruped agents. The composer policy links pair of experts via transitions to a sampled target state, allowing experts to be composed sequentially. Each expert specializes in a single skill, such as a locomotion gait or a jumping motion. Instead of a hierarchical or mixture-of-experts architecture, we train a single composer policy in an independent process that is not conditioned on the other expert policies. By reusing the same composer policy, our approach enables adding new experts without affecting existing ones, enabling incremental repertoire expansion and preserving original motion quality. We measured the transition success rate of 72 transition pairs and achieved an average success rate of 99.99\%, which is over 10\% higher than the baseline random approach, and outperforms other state-of-the-art methods. Using domain randomization during training we ensure a successful transfer to the real world, where we achieve an average transition success rate of 97.22\% (N=360) in our experiments.

% We propose the expert composer policy, a framework to reliably expand the skill repertoire of quadruped agents. Each expert specializes in a single skill, such as a locomotion gait or a jumping motion. \ct{We introduce an independent composer policy module that consolidates all skills into a coherent meta-controller. Instead of a hierarchical or mixture-of-experts architecture, we train a single composer policy in an independent process that is not conditioned on the other expert policies. The composer policy links any pair of experts via transitions to a target state, allowing experts to be composed sequentially. By reusing the same composer policy}, our approach enables adding new experts without changing existing ones, enabling incremental repertoire expansion and preserving original motion quality. \ct{We measured the transition success rate of 72 transition pairs and achieved an average success rate of 99.99\%, which is over 10\% higher than the baseline random approach}, and outperforms other state-of-the-art methods. \ct{Using domain randomization during training we ensure a successful transfer to the real world, where we achieved an average transition success rate of 97.22\% in our experiments}.
\end{abstract}

%%%%%%%%%%%%%%%%%%%%%%%%%%%%%%%%%%%%%%%%%%%%%%%%%%%%%%%%%%%%%%%%%%%%%%%%%%%%%%%%

\section{Introduction}

As the robotics community refines, improves, and develops new skills for robots, an issue arises: how to enable access to a vast repertoire of skills?  Different tasks commonly require different controllers regarding inputs, learning signals, such  model architectures, such that switching from one controller to another can be unstable or impossible. The problem is further magnified in dynamic locomotion controllers. Focusing on real-world deployment, we wish to keep existing skills intact while incorporating new abilities that interconnect with existing ones, creating an ever-expanding library of skills.

A common approach to handle multiple skills is through a mixture of experts and hierarchical controllers.  Low-level experts are trained in a single skill and integrated into a coherent controller using a higher-level policy, through gating modules that mixes actions of low-level experts with additive \cite{won2020scadiver, zhang2018mann}, or multiplicative composition \cite{peng2019mcp}. However, adding a new low-level expert requires re-training the controllers, which inevitably affects the behavior of the low-level experts, degrading their quality and limiting the ability to scale the number of skills robustly \cite{soeseno2021transition}.

\begin{figure}[ht!]
  \centering  
  \includegraphics[width=1.0\columnwidth]{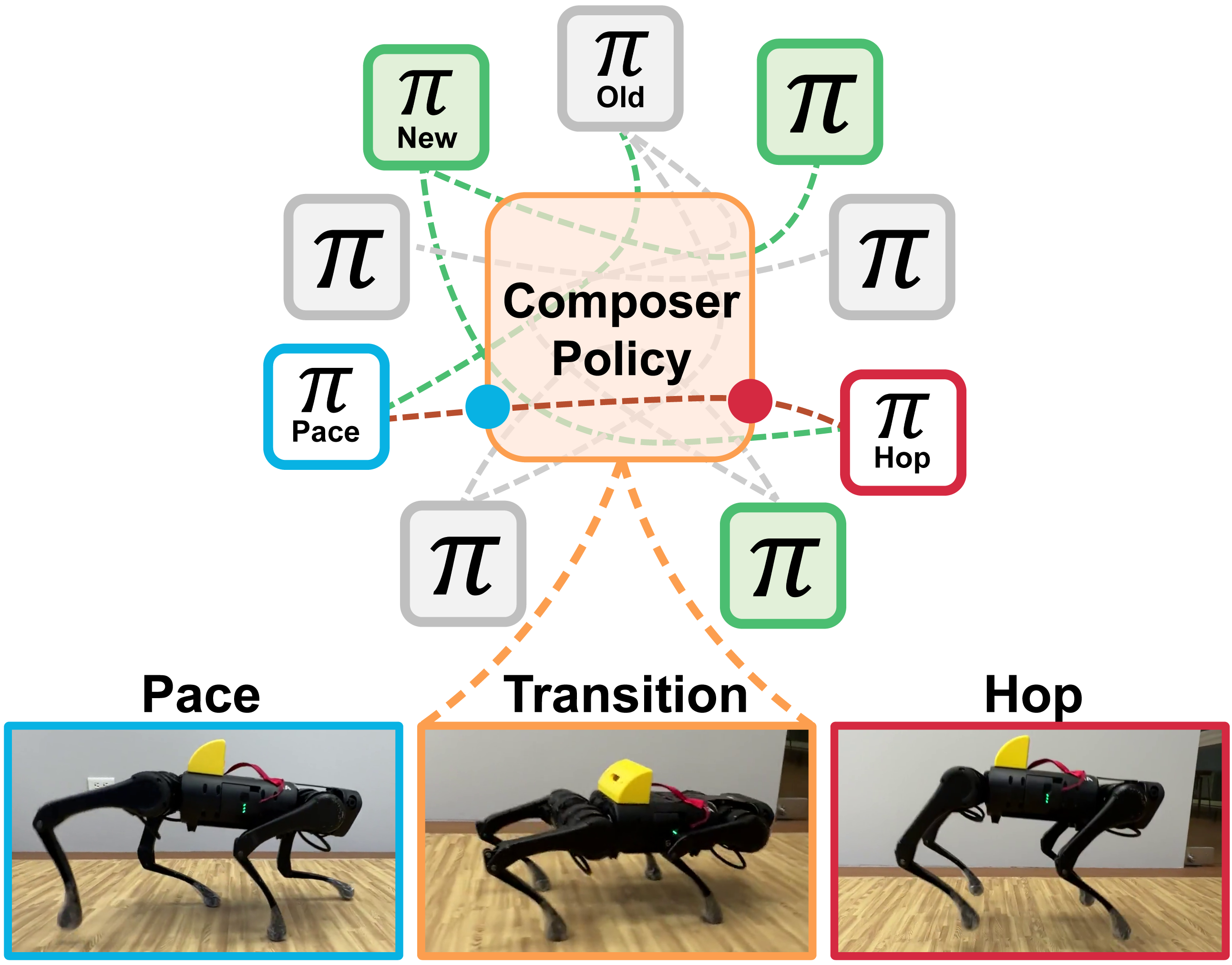}
    \caption{Expert composer policy generates novel transitions between arbitrary agent states, enabling skill repertoire expansion while preserving the motion quality of the original experts.}

    % can incrementally expand the skill repertoire of a character with novel transition motions between skill pairs.
    \label{fig:teaser}
\end{figure}

Rather than mixing the experts, we propose keeping them independent and using a mechanism to compose the skills over time sequentially. This goal of preserving the quality and behavior of low-level experts is prevalent in character controllers used in animation \cite{Kovar2002motion, safonova2007interpolatedgraph}.  A similar premise exists in physics-simulated characters \cite{kangkang2016controlgraph, soeseno2021transition}, where transition motions derive from the careful selection of motion parameters such as timing and agent poses.  However, these approaches execute transitions between controllers through an instantaneous switch mechanism, limiting their applicability for linking skills with distinctive motions.

In this work, we propose a framework that enables a large repertoire of skills, via smooth and dynamic transition trajectories. This is line with works in skill chaining and policy sequencing \cite{lee2018composing, bagaria2019option, lee2021adversarial}. We train individual skills with dedicated policy controllers, further referred to as experts. These experts are trained in a physics-enabled simulation environment with randomized physical properties to enable transfer to real-world robots. Next, we introduce our expert composer policy that generates transition trajectories between arbitrary agent states, bridging every pair of experts smoothly. Inspired by \cite{ma2021spacetimebounds}, we employ tolerance bounds on physical properties that shrink over time to generate sensible trajectories.  With a large skill repertoire of experts and the composer policy, we can create a coherent controller by switching to the composer policy between the execution of two experts (Figure \ref{fig:teaser}). With our method, adding new experts does not require additional training processes, allowing us to maintain all previously learned skills while accommodating new ones. Our list of contributions is as follows:

\begin{itemize}[leftmargin=1em, itemindent=0pt]
    \item A novel transition mechanism that generates smooth trajectories between arbitrary states of physics-enabled agents learned via shrinking boundaries,
    \item A robust process for skill repertoire expansion through the transition mechanism, and
    \item A versatile meta-controller capable of performing various skills that can be incrementally expanded at deployment.
\end{itemize}

\section{Related Works}

\textbf{Traditional Controllers.} For a long time, roboticists have tried to emulate the efficiency and elegance of locomotions in nature \cite{alexander1984gaits, hoyt1981gait}.  A gait is a periodic sequence of foot contacts with the ground that produces locomotion \cite{haynes2006gaits} and is achievable via analytical models of the dynamics of a legged system \cite{wieber2016modeling} or by using central pattern generators (CPG) to produce oscillatory motion patterns and using sensors for close-loop trajectory optimization \cite{fukuoka2013analysis, ijspeert2008central, barasuol2013reactive}. 
Model predictive control (MPC) is a popular type of approach that produces optimized and distinct gaits \cite{farshidian2017real}.  These have been extensively used for quadruped robots, with simplified dynamics and convex optimization \cite{di2018dynamic}, linearizing the dynamics and formulating the problem with quadratic programming \cite{ding2019real}, as well as approaches that use the full dynamics of the system including feet contacts \cite{neunert2018whole}.

\textbf{Learning-Based Controllers.} With deep reinforcement learning (RL), it is possible to train controllers for task-based objectives \cite{sutton2018reinforcement, margolis2023walk}.  Learning-based controllers have been used to tackle challenging terrain using just proprioception \cite{lee2020learning}, as well as with other sensors such as with depth \cite{agarwal2023legged} and with LiDAR \cite{miki2022learning}.  In our work, we learn locomotion policies based on a motion imitation framework \cite{peng2018deepmimic}, where the agent imitates a reference animation while interacting with physics-enabled environments \cite{Mourot2022-je, li2023versatile, vollenweider2023advanced}.  The controllers can be deployed to the real world using methods such as domain randomization \cite{tobin2017domain}  during training or domain adaptation with real-world trajectories \cite{peng2020learning}.

\textbf{Transition between Controllers.} Solving complex tasks requires versatile controllers. A common way to coordinate skills is through a hierarchical architecture of low-level and high-level controllers \cite{peng2019mcp, Jungdam2020}. A task-based objective, in the form of a reward function, drives the behavior of the controllers \cite{peng2021amp, jain2019hierarchical, yang2021fast}. It is also possible to integrate kinematic motion controllers \cite{zhang2018mann, starke2019nsm, starke2020localphase} with other high-level controllers trained to interact with other agents in a physics environment  \cite{bergamin2019drecon, park2019learning}.

To avoid costly re-training and better scalability, we want to compose the controllers sequentially rather than mix them hierarchically. Sequential execution of learned skills is not a recent idea \cite{kober2014movement, pastor2009learning}. Also called skill chaining\cite{bagaria2019option, konidaris2009skill}, it has been used to tackle long-horizon tasks \cite{lee2021adversarial}, splitting a complex task into separate sub policies and coordinating their execution \cite{lee2019learning, clegg2018learning}. Other approaches include discovering transition timing via statistical methods conditioned on motion phases \cite{soeseno2021transition}, using neural nets conditioned on latent states \cite{christmann2023expanding}, and parametrized transitions~\cite{haynes2011gait, boussema2019online}.

Most similar to ours is the work of \cite{lee2018composing}, where each expert has a matching transition policy that drives the agent to a feasible starting state. The policies are trained by rolling out transition episodes, labeling trajectories a success or failure, and optimizing with RL over the expected success rate. In our proposed work, the composer policy is trained completely in isolation from the other experts. Rather than explicitly optimizing for the success rate, it is trained to drive the robot to a target state. Furthermore, we train only a single policy to transition between any pair of experts.

\begin{figure}[t]
    \centering
    \vspace{0.15cm}
    \includegraphics[width=1.0\columnwidth]{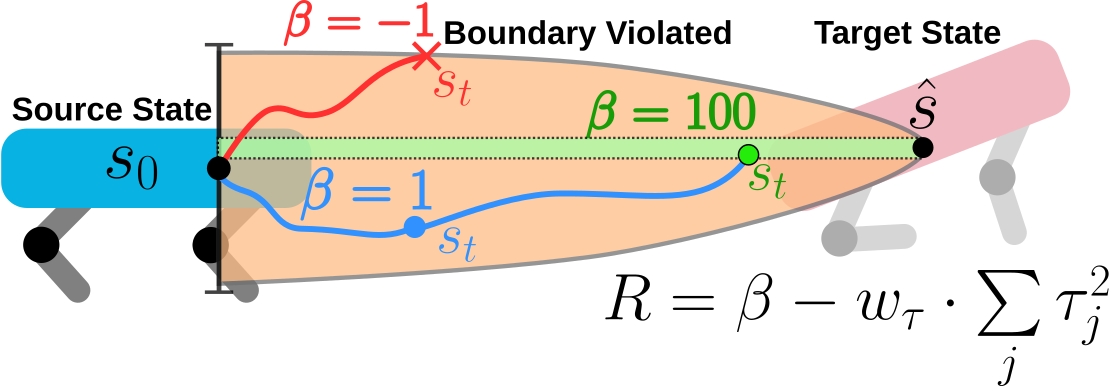}
    \caption{Illustration of physical-property bounds on the agent's states.}
    \label{fig:transition_bounds}
\end{figure}

\section{Method}

% Construct a library of robust, independent locomotion experts that can be deployed in the real-world. The robot can execute any individual expert but cannot connect them together for more complex motion patterns.
We begin by training independent controllers. These single-skill experts are physics-based controllers trained via motion imitation \cite{peng2018deepmimic, peng2020learning}. Then, we introduce the composer policy $\mathbf{P}$, a module that generates trajectories to arbitrary agent states, allowing the agent to transition between learned skills freely.  $\mathbf{P}$ learns to drive the agent from a starting state to a target state sampled from the distribution of the target skill. Expansion is done by simply adding a new expert and using $\mathbf{P}$ to mediate transitions with the existing experts.  

% Figure~\ref{fig:schematic-diagram} depicts the entire process of our method, as well as the composer policy $\mathbf{P}$ as a diagram of interconnected experts.

\subsection{Experts Initialization}
\label{sec:method-experts-initialization}
Each expert is a reinforcement learning policy trained by tracking a reference motion clip in a physics simulator with a motion imitation framework. Suppose we have $k$ number of skills, each described by a reference motion clip.  Each expert is denoted as a policy $\pi_i(\bm{a}|\bm{s}, \bm{g}_i);\ i \in 1...k$ that outputs the actions $\bm{a}$, executed with a PD controller. The actions are conditioned on the current state of the character $\bm{s}_t$ as well as data from the reference motion clip $\bm{g}_i$ from four future frames \cite{peng2020learning}.  Each expert policy trains on a single reference motion clip.  Our reward function design is similar to existing motion imitation frameworks \cite{peng2018deepmimic}.

We employ domain randomization extensively to ensure each expert is robust enough to be deployed with the real-world robot.  Specifically, we randomize the mass of each link in the robot, introduce disturbance forces, add noise to sensor readings, and randomize the terrain height and friction.  The details of the training configuration for each expert are further discussed in Section \ref{sec:implementation-details}.

% To ensure the robustness of each expert for deployment on the real-world robot, we employ domain randomization extensively. This includes randomizing the mass of each robot link, introducing random disturbance forces, adding noise to sensor readings, and randomizing terrain height and friction. Further details regarding the training configuration for each expert are discussed in Section 5.

\subsection{Composer Policy \texorpdfstring{$\mathbf{P}$}{P} for Novel Transitions}
\label{sec:method-pc}

% With experts that independently accommodate each skill, we now focus on coherently composing them into a single controller that can perform the entire skill repertoire. In a Mixture-of-Experts setup, this is achieved by training a gating network that uses either an additive \citep{won2020scadiver, zhang2018mann} or multiplicative \citep{peng2019mcp, luo2020carl} process to blend the actions of the experts. However, to add new skills, the gating network needs to be re-trained entirely or fine-tuned to accommodate the new experts, which leads to quality degradation of the experts. To avoid re-training, one can search for the appropriate timing to perform an instant switch between a source and a target expert such as in the TMT setup \citep{soeseno2021transition}. However, it cannot handle transitions between experts that are too distinct.

% \begin{figure*}[ht]
%   \centering
%   \includegraphics[width=0.85\textwidth]{Figures/schematic-diagram.png}
%   \caption{Method overview, from the initialization of experts through imitation learning and domain randomization, followed by independent training of the composer policy to mediate transitions. The result is a coherent meta-controller that connects experts through novel transitions.}
%   \label{fig:schematic-diagram}
% \end{figure*}

\begin{figure*}
    \centering
    \vspace{0.15cm}
    \includegraphics[width=0.99\textwidth]{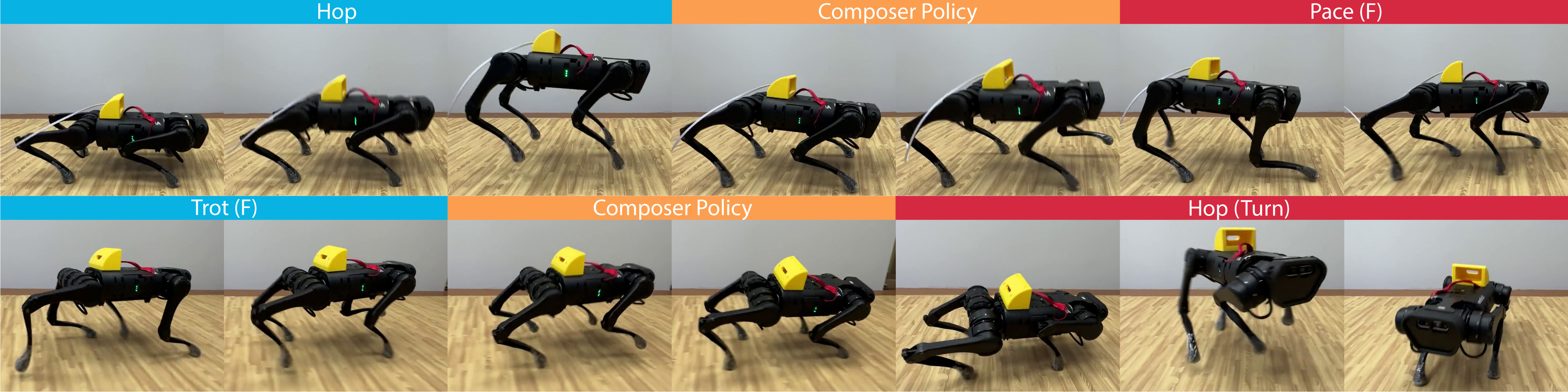}
    \caption{A visualization of the composer policy with the real-world robot. Please watch the supplementary video for more extensive demonstrations.}
    \label{fig:real-world-demo}
\end{figure*}

The composer policy $\mathbf{P}$ is an intermediate policy that drives the agent to a target state by generating novel transition trajectories. $\mathbf{P}(\bm{a}_t|\bm{s}_t, \hat{\bm{s}})$ takes in the current state of the character $\bm{s}_t$, and a target state $\hat{\bm{s}}$. The state $\bm{s}_t$ is the same as the experts and consists of position, orientation, binary feet contact indicators, linear and angular velocities, and joint angles and velocities.  The fixed target state $\hat{\bm{s}}$ contains the same information as $\bm{s}_t$, minus the feet indicators,  and represents the state of the agent should be after the transition.

$\mathbf{P}$ is trained with episodic reinforcement learning in a completely independent process from the other experts. During training, we sample the source and target states (with added randomizations) directly from the animation data. During each episode, the goal of $\mathbf{P}$ is to match the target state $\hat{\bm{s}}$. Potential-based rewards \cite{badnava2023new} could be used to incentivize the agent to match the properties of the target state. However, in early experiments, we found it difficult to tune the reward elements to ensure the agent matched \emph{all} properties of the target state. Instead, we train $\mathbf{P}$ with a simple indicator function based on a set of narrowing hard boundaries with only a single penalization term for energy efficiency.

% Potential-based reward shaping can be time-consuming, difficult to tune and not generalize well to different transition pairs. Instead, we define constraints around the desired target state and let the agent discover trajectories within the defined boundaries.

% The boundaries define a tolerance region which starts large and is annealed to a narrow space around the target state. It guides the agent without over-constraining the intermediate trajectory.
% Transition duration is dynamic, with early finish conditions. Agent is incentivized to learn to finish early with bonus reward function.
The reward function incentivizes the agent to stay within a set of boundaries for each relevant property: joint positions, linear and angular velocity, and orientation \cite{ma2021spacetimebounds}. It establishes tolerance regions where the agent can remain and receive a reward of $1$.  When the agent violates a boundary, the punishment is a negative reward and termination of the episode.  If the agent remains within the tolerances until the end of the episode, it receives a large positive reward.  Critically, the boundaries are annealed around a line that connects the initial transition state to the target state.  The agent learns to stay clear of the shrinking boundaries while closing into the target state (Figure~\ref{fig:transition_bounds}). The transition duration is dynamic, and an episode finishes early if all states match the final tolerances. In fact, we expect and observe that most transitions do terminate early if $\mathbf{P}$ is working well. The boundary indicator function has the form

% To incentivize the composer policy to generate motion trajectories that have desirable properties such as smoothness, as well as obeying certain physical limitations, we impose training constraints in the form of tolerance bounds inspired by \citep{ma2021spacetimebounds}. \ysluo{During training, the policy aims to reach target state as fast as possible, while performing actions that maintain the state of the agent within a computed tolerance, the feasible boundary. The boundaries establish tolerance regions, initially starting with a wide range around the source state and are annealed to narrow spaces around the target state for each relevant physical property, see Figure \ref{fig:bounds-illustration}. We employ early termination conditions where the episode is terminated when either all physics properties of the agent are within the tolerance or if any of the physics properties are violated. The tolerance bound is defined as,}

\begin{equation}
    \label{eq:feasible-bounds}
    \beta = 
    \begin{cases}
        100,  & \text{if } \text{max}(|\bm{s}_t - \hat{\bm{s}}| - \bm{\sigma}_e) \le 0  \\
        -1,  & \text{if } \text{max}(|\bm{s}_t - \Psi(\bm{s}_0, \hat{\bm{s}}, t)| - \bm{\sigma}_t) \ge 0 \\
        1, & \text{otherwise}
    \end{cases},
\end{equation}

\noindent where $\Psi(\bm{s}_0, \hat{\bm{s}}, t)$ is a linear function that connects the initial state of the agent $\bm{s}_0$ to the target state $\hat{\bm{s}}$, and the tolerance $\bm{\sigma}_t$ is annealed at each time step $t$ according to the equation

\begin{equation}
    \bm{\sigma}_t = \bm{\sigma}_s + t^p \ (\bm{\sigma}_e-\bm{\sigma}_s),
\end{equation}

% Description of the reward shaping, tolerance bounds equation, extra penalization terms for regularization (torque).
\noindent with $\bm{\sigma}_s$ and $\bm{\sigma}_e$ denoting the tolerance at the start and at the end of the transition period, and $p$ is an exponential parameter to modify how the boundary shrinks (Table \ref{tab:pc-params}).  With boundary annealing, the composer policy learns to match the target state closely.  However, it does not guarantee that the trajectories are smooth or energy efficient. For this, we introduce a penalization term for joint torques. The complete reward function is as follows, where $\tau_j$ is the torque of $j$-th joint of the agent and $w_\tau$ is a scalar that controls the scale of penalization,

\begin{equation}
    {
    R = \beta - w_{\tau}\sum_j{\tau_j^2}.
    }
\end{equation}
% \noindent where $c \in \{0, 1\}$ is a binary variable that is set to $c=1$ if any of the end-effectors of the agent is touching the ground, and $c=0$ otherwise. In addition, $w_c$ and $w_\tau$ are scalars that control the severity of the penalization terms; and $\tau_j$ is the torque of $j$-th joint of the agent.

\subsection{Composition of Experts}
\label{sec:method-coe}

\begin{figure}[b!]
    \centering
    \includegraphics[width=1.0\columnwidth]{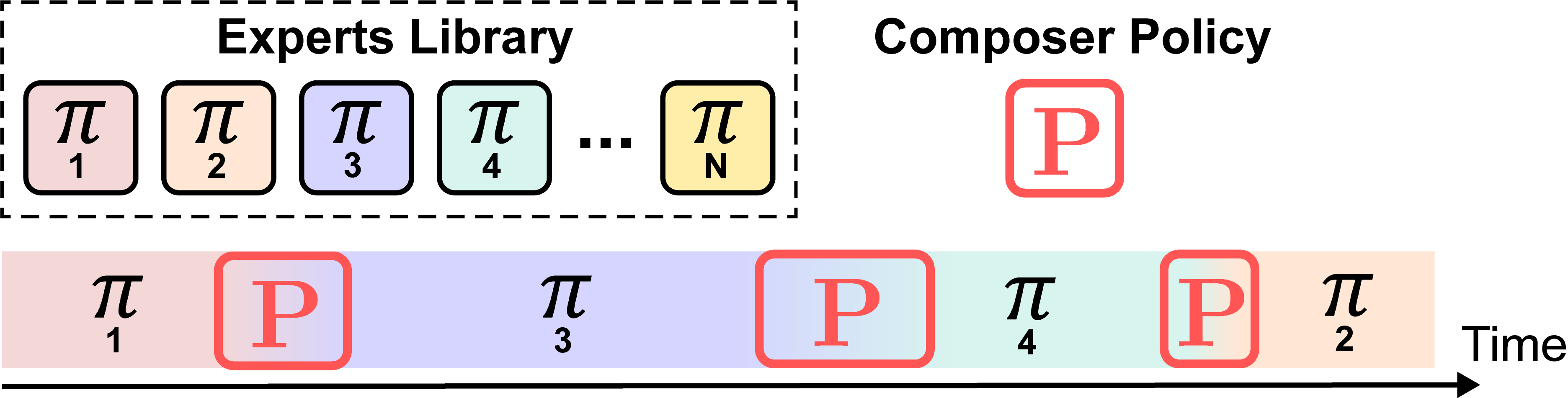}
    \caption{Illustration of the composer policy $\mathbf{P}$ enables the sequencing of any expert policy from the library over time.}
    \label{fig:transition_timeline}
\end{figure}

After training the composer policy $\mathbf{P}$, we can sequentially compose the experts (Figure \ref{fig:transition_timeline}). The agent may start executing a pace skill with $\pi_{\text{pace}}$. At some later point, an event triggers a switch to a different expert, such as $\pi_{\text{hop}}$. To execute this transition, $\mathbf{P}$ takes control of the agent and samples a target state $\hat{\mathbf{s}}$ from within the distribution of $\pi_{\text{hop}}$, using its animation data. Then, $\mathbf{P}$ performs actions such that the character's state at the end of the transition is close enough to the target state that the new expert can take over control. The transition ends with the target policy $\pi_{\text{hop}}$ taking over control of the character. This process can be repeated indefinitely and robustly for any pair in the policy library.

If $\mathbf{P}$ is trained with a large enough distribution of start and target states, it is reasonable to assume that it should learn feasible transitions between any experts, even new ones.  This is possible because $\mathbf{P}$ only requires a target state and does not directly depend on or use information from the experts in the library. Therefore, the expansion can be done incrementally and indefinitely by adding new experts and using the same $\mathbf{P}$, without re-training or fine-tuning.

% \invalid{[MOVE TO 3.3?]To guarantee the scalability of the system, the composer policy only needs to be trained once and independently of the experts in the skill repertoire. The same composer policy is used when the repertoire is expanded. This is achieved through a pair-wise transition from a source $\pi_m$ to the target $\pi_n$ through visiting the composer policy $\mathbf{P}$ as illustrated in Figure \ref{fig:schematic-diagram}.}
\section{Implementation Details}
\label{sec:implementation-details}
% This section describes the necessary implementation details to reproduce our experiments' results, including adapting each expert to real-world robots through imitation learning with domain randomization and training the composer policy to generate transition trajectories.

\subsection{Hardware and Locomotion Experts}
Our agent is a Unitree A1 quadruped robot with 12 actuators.  We implement a motion imitation framework \cite{peng2018deepmimic, peng2020learning} with Isaac Gym \cite{makoviychuk2021isaac} and use a consumer-grade laptop equipped with an Intel 8-core i7-11800H 2.3 GHz and an NVIDIA RTX 3070 8GB with the PPO-clip loss \cite{schulman2017ppo} to train the experts. 
To transfer the policies from simulation to the real world, we employ domain randomization \cite{Margolis-RSS-22, kumar2021rma} with parameters shown in Table~\ref{tab:dr-params}.  The agent states include linear velocity estimated in the real world by fusing IMU readings and the leg velocity during feet contacts. The locomotion experts are deployed in a zero-shot manner and can execute for long periods without failures.

\subsection{Details of the Composer Policy}
The architecture of the composer policy $\mathbf{P}(\bm{a}_t|\bm{s}_t, \hat{\bm{s}})$ is a 2-layer feed-forward neural network with 512 and 256 hidden units.  Each layer uses ELU activations except for the linear output layer.  The composer policy receives an observation vector of 208 dimensions which consists of the three latest states of the agent, the actions from the past three timesteps, the target state $\hat{\bm{s}_t}$, and the centers of the tolerance boundaries at the current timestep computed from $\Psi(\bm{s}_0, \hat{\bm{s}}, t)$.  The observations also include a single scalar that encodes the normalized time, starting at 0.0 and increasing to 1.0 at the end of the transition period (episode).  The state $\bm{s}_t$ consists of position, orientation, binary feet contact indicators, linear and angular velocities, and joint angles and velocities.  The target state $\hat{\bm{s}}$ is the same, minus the feet indicators.  The maximum episode length is set to 2 seconds.  The composer policy outputs 12 target joint angles, actuated using a PD-controller.  Table~\ref{tab:pc-params} contains detailed parameters for the tolerance boundaries and reward penalization terms.

\begin{table}[t!]
    \centering
    \vspace{0.2cm}
    \caption{Parameters for domain randomization. Ranges are sampled uniformly. The feet contacts value indicates the probability of zeroing out the contacts, per foot.}
    \label{tab:dr-params}

    % \begin{tabular}{ |l|c|c| } 
    %     \hline
    %     \textbf{Parameter} & \textbf{Value} & \textbf{Type} \\ \hline
    %     Action Noise & 0.02 & Additive \\ 
    %     Rigid Bodies Mass & [0.95, 1.05] & Scaling \\
    %     Stiffness Gain (PD Controller) & [29, 41] & -- \\
    %     Damping Gain (PD Controller) & [0.9, 1.5] & -- \\
    %     Sensor Noise - Orientation & 0.06 & Additive \\
    %     Sensor Noise - Linear Velocity & 0.15 & Additive \\
    %     Sensor Noise - Angular Velocity & 0.06 & Additive \\
    %     Sensor Noise - Joint Angles & 0.02 & Additive \\
    %     Sensor Noise - Joint Velocity & 1.5 & Additive \\
    %     Sensor Noise - Feet Contacts & 0.1 & Probability \\
    %     \hline
    % \end{tabular}

    \begin{tabular}{ |l|c|c| } 
        \hline
        \rule{0pt}{10pt}\multirow{2}{*}{\textbf{Parameter}} &
        \multicolumn{2}{c|}{\textbf{Value}} \\ 
        \cline{2-3}
        & \rule{0pt}{10pt} \textbf{Experts} & \textbf{Composer Policy} \\
        \hline
        Action Noise                    & $\pm 0.02$ & $\pm 0.02$ \\ 
        Rigid Bodies Mass               & [75\%, 125\%] & [95\%, 105\%] \\
        P Gain (PD Controller)  & [35, 65] & [45, 55] \\
        D Gain (PD Controller)    & [1.0, 1.4] & [0.9, 1.2] \\
        Ground Friction                 & [0.1, 1.5] & [0.1, 1.5] \\
        Noise - Orientation      & $\pm 0.05$ & $\pm 0.06$ \\
        Noise - Linear Velocity  & $\pm 0.25$ & $\pm 0.25$ \\
        Noise - Angular Velocity & $\pm 0.3$ & $\pm 0.3$ \\
        Noise - Joint Angles     & $\pm 0.02$ & $\pm 0.02$ \\
        Noise - Joint Velocity   & -- & $\pm 1.5$ \\
        Noise - Feet Contacts    & 20\% & 20\% \\
        \hline
    \end{tabular}
\end{table}
\begin{table}[t!]
    \centering
    \caption{Tunable Parameters of the composer policy. $\sigma_s$ and $\sigma_e$ indicate the tolerance at the start and end of the transition, respectively.}
    \label{tab:pc-params}
    \begin{tabular}{ |l|c|c|c| } 
        \hline
        \textbf{Component}& \textbf{Value} & $\sigma_s$ & $\sigma_e$ \\ \hline
        CoM - Height & $p = 2$ & 0.35 & 0.02 \\ 
        Orientation & $p = 4$ & 1 & 0.2 \\ 
        Linear Velocity & $p = 8$ & 2.5 & 0.2 \\ 
        Angular Velocity & $p = 8$ & 15 & 0.2 \\ 
        Joint Angles & $p = 2$ & 3.14 & 0.5 \\ 
        % Contact Term & $w_c = 0.01$ & -- & -- \\
        Torque Term & $w_\tau = 0.0001$ & -- & -- \\
        \hline
    \end{tabular}
\end{table}
\section{Experiments and Results}
\label{sec:experiments}

\begin{figure*}[ht!]
  \centering
  \vspace{0.15cm}
  \includegraphics[width=1.0\textwidth]{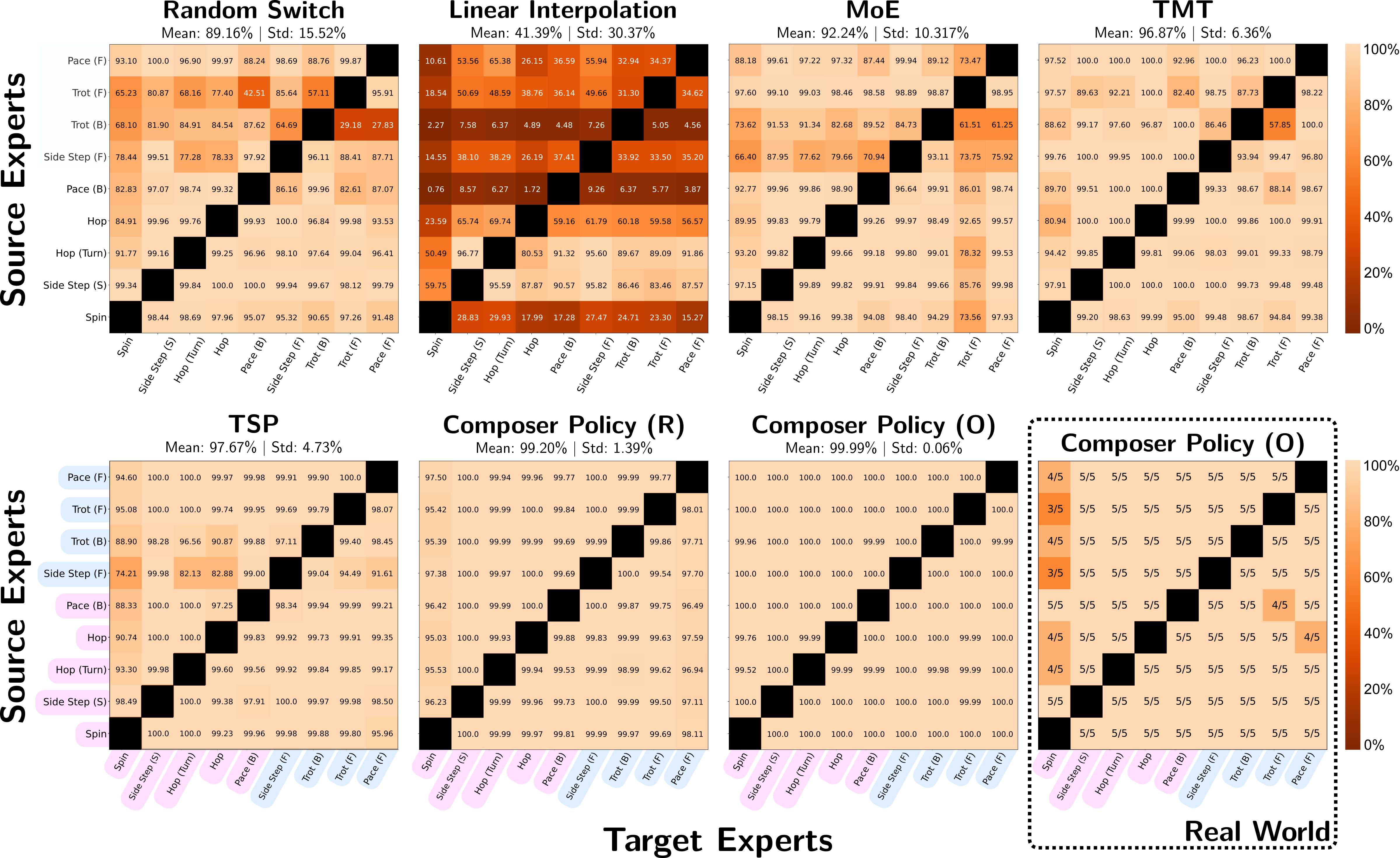}
  \caption{Success rates of the composition strategies for 72 transition pairs. (F) and (B) indicate forward and backward for Trot and Pace, (F)ast and (S)low for Side Step. Experts highlighted in blue were part of the training set (N=4), and purple indicates new experts added to the library after training (N=5). Our \textbf{Composer Policy (R)} and \textbf{(O)} variants outperforms the baselines and existing approaches in simulation by a significant margin. It is also successful in the real-world, with just 10 failures out of 360 trials.}
  \label{fig:transition-success-rates}
\end{figure*}

We constructed a library of 9 distinct experts in the following motions: \textit{Trot (F)}, \textit{Trot (B)}, \textit{Pace (F)}, \textit{Pace (B)}, \textit{Hop}, \textit{Hop (Turn)}, \textit{Spin}, \textit{Side Step (F)ast}, and \textit{Side Step (S)low}. This results in a total of 72 unique transition pairs. To evaluate generalization to new skills, $\mathbf{P}$ was trained by sampling starting and target states from just 4 experts: \textit{Trot (F)} and \textit{(B)}, \textit{Pace (F)} and \textit{Side Step (F)}. The 5 remaining experts were added in a supposed expansion process, i.e. were not part of the training. We evaluated the composer policy with two target sampling methods. \textbf{Composer Policy~(R)} samples the target from a random phase of the animation of the target expert. \textbf{Composer Policy~(O)} samples from an optimal phase interval after analysis of the success rate of \textbf{(R)}, detailed in Section~\ref{sec:exp-transition-analysis}.We compare our approach to two naive baselines and three existing methods from the literature:

\begin{itemize}[leftmargin=1em, itemindent=0pt]
    \item \textbf{Random Switch} -- Instantaneously switches control from the source to the target expert at a random time.
    \item \textbf{Linear Interpolation} -- Linearly interpolates the target joint angles from the start to the target state for 0.5 seconds.
    \item \textbf{Mixture-of-Experts (MoE)} \cite{won2020scadiver}  -- A gating network is trained to output a 9-dimensional vector of activation weights for each expert. \textbf{MoE} was trained with access to all 9 experts. A new target expert was randomly selected every 3 seconds. The policy was rewarded for matching the animation of the selected expert.
    \item \textbf{Transition Motion Tensor (TMT)} \cite{soeseno2021transition} -- The TMT performs an instantaneous switch to the target expert at an optimal timing. The timing is determined offline by a Monte Carlo approach to determine the source and target phases with the highest success rate.
    \item \textbf{Target State Proximity (TSP)} \cite{lee2018composing} -- A policy with a reward function in the style proposed by \cite{lee2018composing}: $R = P(s_{t+1}) - P(s_t)$. Originally, the proximity function $P(s)$ is modeled by a neural-net, but, because we have access to the animation of the target expert, we sample it directly and compute the euclidean distance. The \textbf{TSP} is trained and evaluated under the same settings as our \textbf{Composer Policy}. However, \textbf{TSP} requires one transition policy per expert. Based on motion similarity, the model \textit{Trot (B)} was used for the expansion skill \textit{Pace (B)}, \textit{Side Step (F)} for \textit{Side Step (S)}, and \textit{Pace (F)} for the remaining expansion skills.
\end{itemize}

% Section~\ref{sec:exp-success-rates} presents the results collected in simulation, Section~\ref{sec:exp-real-world} presents the results from the real-world trials, and Section~\ref{sec:exp-transition-analysis} provides additional analysis of the transition and the behavior of the composer policy $\mathbf{P}$.
\subsection{Success Rate in Simulation}
\label{sec:exp-success-rates}

\begin{figure}[!ht]
    \centering
    \vspace{0.15cm}
    \includegraphics[width=0.9\columnwidth]{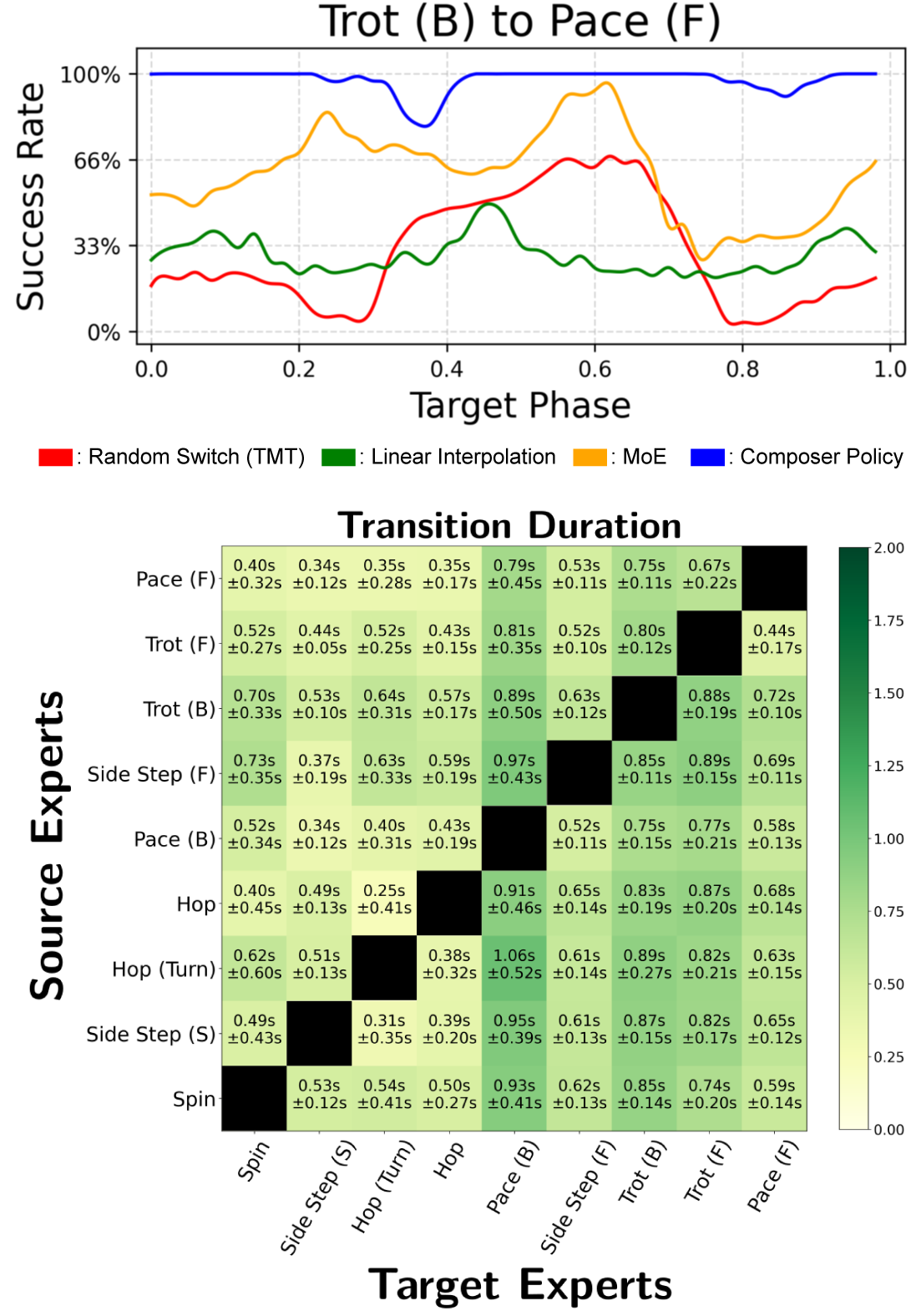}
    \caption{Success rate across all target phases (top) and transition duration of our composer policy (bottom). By excluding regions of lower success we sample targets from an optimal interval.}
    \label{fig:transition_analysis}
\end{figure}

For each strategy, an evaluation episode consisted of three stages: pre-transition, transition, and post-transition. In the pre-transition stage, the source expert is executed. It is followed by the transition stage where one of the methods takes control of the agent. Finally, the target expert takes control for 10 seconds in the post-transition stage. A transition is considered a failure if any links other than the feet of the agent touch the ground during the transition or post-transition stages. For each pair and method we executed 50k evaluation episodes, the results of which are consolidated in Figure~\ref{fig:transition-success-rates}.

\begin{figure}[ht]
  \centering
  \vspace{0.15cm}
  \includegraphics[width=0.95\columnwidth]{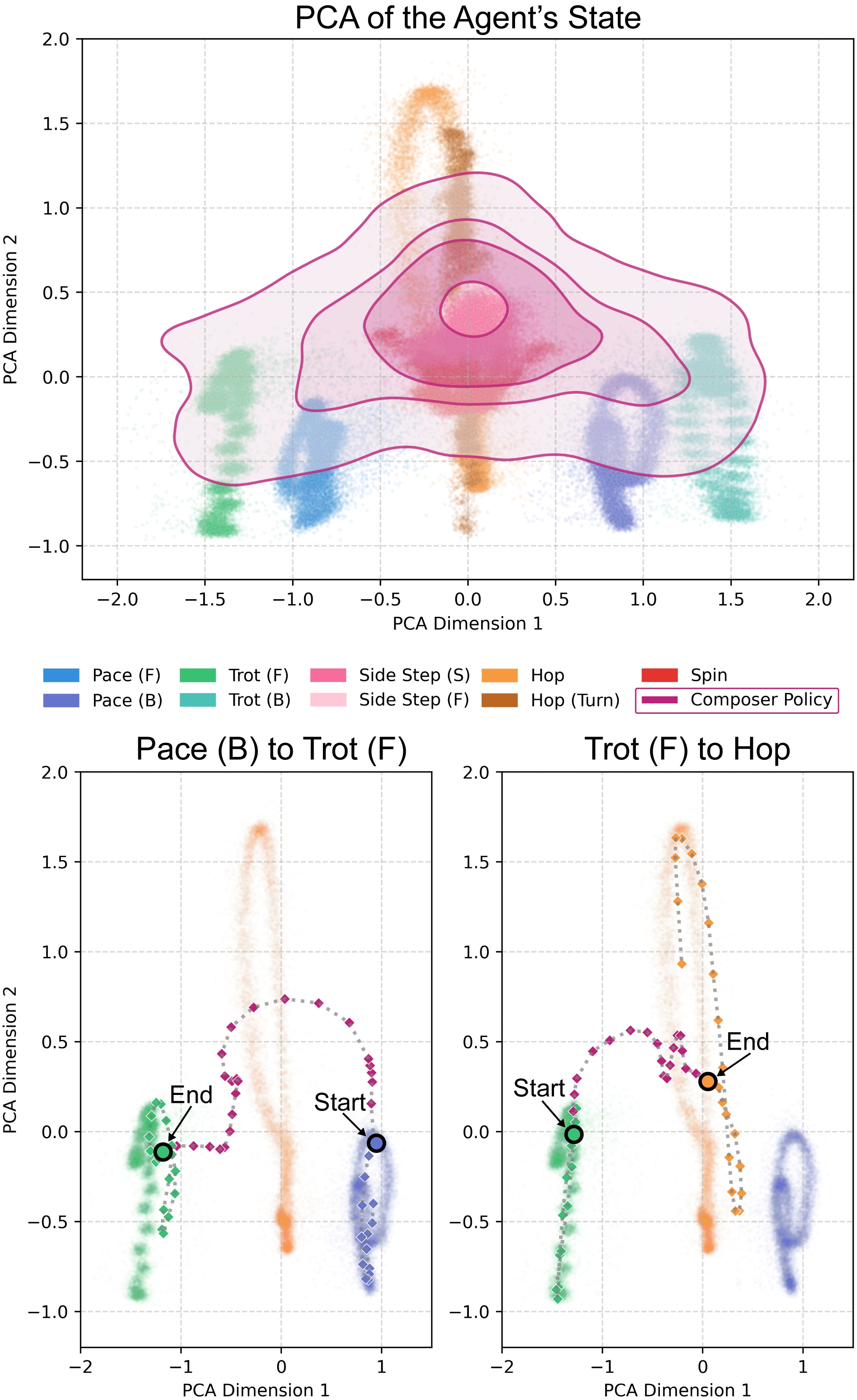}
  \caption{PCA visualization of the experts distribution (top) and transition trajectories (bottom). The purple contours depicts the density distribution of the Composer Policy's states, where the outermost includes 95\% of the data. The bottom trajectories show that our method can drive the agent's state between the different distributions effectively.}
  \label{fig:pca-and-trajectories}
\end{figure}

The baseline \textbf{Random Switch} provides some information on the difficulty of the transition pairs. We can observe that the lowest success occurred for pairs with distinct motions, e.g. moving from forward to backward, and sideways. The \textbf{Linear Interpolation} baseline performed the worst with an average success under 50\%. \textbf{MoE} beat the baselines in challenging motions, but the low success rate combined with degradation of motion quality is undesirable. \textbf{TMT} uses a simple transition mechanism and managed to achieve a high success rate, but performed poorly for a few pairs. \textbf{TSP} was the second-best performing method, with poor performance in few pairs, but specially when the target expert was the out-of-distribution \textit{Spin} motion. Our approach \textbf{Composer Policy} outperformed the other comparison methods. The random sampling mode \textbf{(R)}, shows that the composer policy can accommodate well the complete range of source and target phases. The performance can be further improved by picking only the best target phases, as demonstrated by the almost perfect success rate of the \textbf{(O)}ptimal sampling method.

\subsection{Success Rate of the Composer Policy in the Real-World}
\label{sec:exp-real-world}

We deployed the same library of experts and composer policy used in the simulation setting to a real-world A1 quadruped robot. For each pair available in the library we executed 5 trials. With 72 available pairs, this resulted in total of 360 trials. A real-world trial consisted of the same three stages of the simulation. However, due to limitations of our physical space, the post-transition stage was limited to 4 seconds. The transition stage was triggered randomly between 2 to 3 seconds after the pre-transition stage. The real-world results are also shown in Figure~\ref{fig:transition-success-rates} (bottom right). We can observe that the simulation performance translated well to the real world, with only 10 failures out of 360 trials, resulting in an average success rate of $97.22\%$. Figure~\ref{fig:real-world-demo} shows deployment of a transitions that starts airborne and into a forward moving motion. 

\subsection{Transition Analysis}
\label{sec:exp-transition-analysis}
We measure the transition success rates for a difficult motion pair across target phases in Figure~\ref{fig:transition_analysis} (top), showing the performance of the \textbf{Composer Policy (R)} for every target phase. By excluding intervals where the success rate is lower than desired, i.e. less than $90\%$ the \textbf{(O)}ptimal sampling method is obtained.  Figure~\ref{fig:transition_analysis} (bottom) shows that transitions between experts moving in the same direction are shorter than for opposing motions. Note that none approaches the 2-second limit, which means the composer policy can guide the agent effectively toward the target state for all tested pairs. Next, we demonstrate the robust generalization capability of the composer policy with a PCA plot of the agent's state in Figure~\ref{fig:pca-and-trajectories} (top). The outermost contour contains 95\% of the samples. We can observe that the distribution of the composer policy is stretched to overlap the majority of the experts' distribution. As long as there is an overlap between the experts and composer policy distribution we should be able to execute a transition. We also provide two examples of transition trajectories, demonstrating the composer policy's capacity to take the state of the agent from one expert to another (Figure~\ref{fig:pca-and-trajectories} bottom).

\section{Discussion} 
\label{sec:conclusion}
We present an approach that allows incremental expansion of a skill repertoire without re-training.  Rather than hierarchically mixing the experts, we train a composer policy to transition between any pairs of experts.  Our method outperforms existing approaches in simulation, achieves a high success rate in the real world, and generates smooth transitions with short duration. 

Our method assumes that the experts are periodic and have an easily accessible distribution of target states that we can sample from. We show that it works well for a large distribution, however, it might not apply to all types of policies, such as perception-based controllers. Overcoming these limitations are critical for building a diverse and expansible skill repertoire for any type of skills.

\bibliographystyle{IEEEtran}
\bibliography{references}

\begin{thebibliography}{10}
\providecommand{\url}[1]{#1}
\csname url@samestyle\endcsname
\providecommand{\newblock}{\relax}
\providecommand{\bibinfo}[2]{#2}
\providecommand{\BIBentrySTDinterwordspacing}{\spaceskip=0pt\relax}
\providecommand{\BIBentryALTinterwordstretchfactor}{4}
\providecommand{\BIBentryALTinterwordspacing}{\spaceskip=\fontdimen2\font plus
\BIBentryALTinterwordstretchfactor\fontdimen3\font minus
  \fontdimen4\font\relax}
\providecommand{\BIBforeignlanguage}[2]{{%
\expandafter\ifx\csname l@#1\endcsname\relax
\typeout{** WARNING: IEEEtran.bst: No hyphenation pattern has been}%
\typeout{** loaded for the language `#1'. Using the pattern for}%
\typeout{** the default language instead.}%
\else
\language=\csname l@#1\endcsname
\fi
#2}}
\providecommand{\BIBdecl}{\relax}
\BIBdecl

\bibitem{won2020scadiver}
\BIBentryALTinterwordspacing
J.~Won, D.~Gopinath, and J.~Hodgins, ``A scalable approach to control diverse
  behaviors for physically simulated characters,'' \emph{ACM Trans. Graph.},
  vol.~39, no.~4, Jul. 2020. [Online]. Available:
  \url{https://doi.org/10.1145/3386569.3392381}
\BIBentrySTDinterwordspacing

\bibitem{zhang2018mann}
H.~Zhang, S.~Starke, T.~Komura, and J.~Saito, ``Mode-adaptive neural networks
  for quadruped motion control,'' \emph{ACM Trans. Graph.}, vol.~37, no.~4,
  Jul. 2018.

\bibitem{peng2019mcp}
X.~B. Peng, M.~Chang, G.~Zhang, P.~Abbeel, and S.~Levine, ``Mcp: Learning
  composable hierarchical control with multiplicative compositional policies,''
  in \emph{NeurIPS}, 2019.

\bibitem{soeseno2021transition}
J.~H. Soeseno, Y.-S. Luo, T.~P.-C. Chen, and W.-C. Chen, ``Transition motion
  tensor: A data-driven approach for versatile and controllable agents in
  physically simulated environments,'' in \emph{SIGGRAPH Asia 2021 Technical
  Communications}, 2021, pp. 1--4.

\bibitem{Kovar2002motion}
\BIBentryALTinterwordspacing
L.~Kovar, M.~Gleicher, and F.~Pighin, ``Motion graphs,'' \emph{ACM Trans.
  Graph.}, vol.~21, no.~3, p. 473–482, Jul. 2002. [Online]. Available:
  \url{https://doi.org/10.1145/566654.566605}
\BIBentrySTDinterwordspacing

\bibitem{safonova2007interpolatedgraph}
\BIBentryALTinterwordspacing
A.~Safonova and J.~K. Hodgins, ``Construction and optimal search of
  interpolated motion graphs,'' \emph{ACM Trans. Graph.}, vol.~26, no.~3, p.
  106–es, Jul. 2007. [Online]. Available:
  \url{https://doi.org/10.1145/1276377.1276510}
\BIBentrySTDinterwordspacing

\bibitem{kangkang2016controlgraph}
L.~Liu, M.~V.~D. Panne, and K.~Yin, ``Guided learning of control graphs for
  physics-based characters,'' \emph{ACM Trans. Graph.}, vol.~35, no.~3, may
  2016.

\bibitem{lee2018composing}
Y.~Lee, S.-H. Sun, S.~Somasundaram, E.~S. Hu, and J.~J. Lim, ``Composing
  complex skills by learning transition policies,'' in \emph{International
  Conference on Learning Representations}, 2018.

\bibitem{bagaria2019option}
A.~Bagaria and G.~Konidaris, ``Option discovery using deep skill chaining,'' in
  \emph{International Conference on Learning Representations}, 2019.

\bibitem{lee2021adversarial}
Y.~Lee, J.~J. Lim, A.~Anandkumar, and Y.~Zhu, ``Adversarial skill chaining for
  long-horizon robot manipulation via terminal state regularization,''
  \emph{Proceedings of Machine Learning Research}, vol. 164, pp. 406--416,
  2021.

\bibitem{ma2021spacetimebounds}
L.-K. Ma, Z.~Yang, T.~Xin, B.~Guo, and K.~Yin, ``Learning and exploring motor
  skills with spacetime bounds,'' \emph{Computer Graphics Forum}, vol.~40,
  no.~2, 2021.

\bibitem{alexander1984gaits}
R.~M. Alexander, ``The gaits of bipedal and quadrupedal animals,'' \emph{The
  International Journal of Robotics Research}, vol.~3, no.~2, pp. 49--59, 1984.

\bibitem{hoyt1981gait}
D.~F. Hoyt and C.~R. Taylor, ``Gait and the energetics of locomotion in
  horses,'' \emph{Nature}, vol. 292, no. 5820, pp. 239--240, 1981.

\bibitem{haynes2006gaits}
G.~C. Haynes and A.~A. Rizzi, ``Gaits and gait transitions for legged robots,''
  in \emph{Proceedings 2006 IEEE International Conference on Robotics and
  Automation (ICRA)}.\hskip 1em plus 0.5em minus 0.4em\relax IEEE, 2006, pp.
  1117--1122.

\bibitem{wieber2016modeling}
P.-B. Wieber, R.~Tedrake, and S.~Kuindersma, ``Modeling and control of legged
  robots,'' in \emph{Springer handbook of robotics}.\hskip 1em plus 0.5em minus
  0.4em\relax Springer, 2016, pp. 1203--1234.

\bibitem{fukuoka2013analysis}
Y.~Fukuoka, Y.~Habu, and T.~Fukui, ``Analysis of the gait generation principle
  by a simulated quadruped model with a cpg incorporating vestibular
  modulation,'' \emph{Biological cybernetics}, vol. 107, no.~6, pp. 695--710,
  2013.

\bibitem{ijspeert2008central}
A.~J. Ijspeert, ``Central pattern generators for locomotion control in animals
  and robots: a review,'' \emph{Neural networks}, vol.~21, no.~4, pp. 642--653,
  2008.

\bibitem{barasuol2013reactive}
V.~Barasuol, J.~Buchli, C.~Semini, M.~Frigerio, E.~R. De~Pieri, and D.~G.
  Caldwell, ``A reactive controller framework for quadrupedal locomotion on
  challenging terrain,'' in \emph{2013 IEEE International Conference on
  Robotics and Automation}.\hskip 1em plus 0.5em minus 0.4em\relax IEEE, 2013,
  pp. 2554--2561.

\bibitem{farshidian2017real}
F.~Farshidian, E.~Jelavic, A.~Satapathy, M.~Giftthaler, and J.~Buchli,
  ``Real-time motion planning of legged robots: A model predictive control
  approach,'' in \emph{2017 IEEE-RAS 17th International Conference on Humanoid
  Robotics (Humanoids)}.\hskip 1em plus 0.5em minus 0.4em\relax IEEE, 2017, pp.
  577--584.

\bibitem{di2018dynamic}
J.~Di~Carlo, P.~M. Wensing, B.~Katz, G.~Bledt, and S.~Kim, ``Dynamic locomotion
  in the mit cheetah 3 through convex model-predictive control,'' in \emph{2018
  IEEE/RSJ international conference on intelligent robots and systems
  (IROS)}.\hskip 1em plus 0.5em minus 0.4em\relax IEEE, 2018, pp. 1--9.

\bibitem{ding2019real}
Y.~Ding, A.~Pandala, and H.-W. Park, ``Real-time model predictive control for
  versatile dynamic motions in quadrupedal robots,'' in \emph{2019
  International Conference on Robotics and Automation (ICRA)}.\hskip 1em plus
  0.5em minus 0.4em\relax IEEE, 2019, pp. 8484--8490.

\bibitem{neunert2018whole}
M.~Neunert, M.~St{\"a}uble, M.~Giftthaler, C.~D. Bellicoso, J.~Carius,
  C.~Gehring, M.~Hutter, and J.~Buchli, ``Whole-body nonlinear model predictive
  control through contacts for quadrupeds,'' \emph{IEEE Robotics and Automation
  Letters}, vol.~3, no.~3, pp. 1458--1465, 2018.

\bibitem{sutton2018reinforcement}
R.~S. Sutton and A.~G. Barto, \emph{Reinforcement learning: An
  introduction}.\hskip 1em plus 0.5em minus 0.4em\relax MIT press, 2018.

\bibitem{margolis2023walk}
G.~B. Margolis and P.~Agrawal, ``Walk these ways: Tuning robot control for
  generalization with multiplicity of behavior,'' in \emph{Conference on Robot
  Learning}.\hskip 1em plus 0.5em minus 0.4em\relax PMLR, 2023, pp. 22--31.

\bibitem{lee2020learning}
J.~Lee, J.~Hwangbo, L.~Wellhausen, V.~Koltun, and M.~Hutter, ``Learning
  quadrupedal locomotion over challenging terrain,'' \emph{Science robotics},
  vol.~5, no.~47, p. eabc5986, 2020.

\bibitem{agarwal2023legged}
A.~Agarwal, A.~Kumar, J.~Malik, and D.~Pathak, ``Legged locomotion in
  challenging terrains using egocentric vision,'' in \emph{Conference on Robot
  Learning}.\hskip 1em plus 0.5em minus 0.4em\relax PMLR, 2023, pp. 403--415.

\bibitem{miki2022learning}
T.~Miki, J.~Lee, J.~Hwangbo, L.~Wellhausen, V.~Koltun, and M.~Hutter,
  ``Learning robust perceptive locomotion for quadrupedal robots in the wild,''
  \emph{Science Robotics}, vol.~7, no.~62, p. eabk2822, 2022.

\bibitem{peng2018deepmimic}
X.~B. Peng, P.~Abbeel, S.~Levine, and M.~van~de Panne, ``Deepmimic:
  Example-guided deep reinforcement learning of physics-based character
  skills,'' \emph{ACM Trans. Graph.}, vol.~37, no.~4, Jul. 2018.

\bibitem{Mourot2022-je}
\BIBentryALTinterwordspacing
L.~Mourot, L.~Hoyet, F.~Le~Clerc, F.~Schnitzler, and P.~Hellier,
  ``\BIBforeignlanguage{en}{A survey on deep learning for skeleton‐based
  human animation},'' \emph{\BIBforeignlanguage{en}{Comput. Graph. Forum}},
  vol.~41, no.~1, pp. 122--157, Feb. 2022. [Online]. Available:
  \url{https://onlinelibrary.wiley.com/doi/10.1111/cgf.14426}
\BIBentrySTDinterwordspacing

\bibitem{li2023versatile}
C.~Li, S.~Blaes, P.~Kolev, M.~Vlastelica, J.~Frey, and G.~Martius, ``Versatile
  skill control via self-supervised adversarial imitation of unlabeled mixed
  motions,'' in \emph{2023 IEEE International Conference on Robotics and
  Automation (ICRA)}.\hskip 1em plus 0.5em minus 0.4em\relax IEEE, 2023, pp.
  2944--2950.

\bibitem{vollenweider2023advanced}
E.~Vollenweider, M.~Bjelonic, V.~Klemm, N.~Rudin, J.~Lee, and M.~Hutter,
  ``Advanced skills through multiple adversarial motion priors in reinforcement
  learning,'' in \emph{2023 IEEE International Conference on Robotics and
  Automation (ICRA)}.\hskip 1em plus 0.5em minus 0.4em\relax IEEE, 2023, pp.
  5120--5126.

\bibitem{tobin2017domain}
J.~Tobin, R.~Fong, A.~Ray, J.~Schneider, W.~Zaremba, and P.~Abbeel, ``Domain
  randomization for transferring deep neural networks from simulation to the
  real world,'' in \emph{2017 IEEE/RSJ international conference on intelligent
  robots and systems (IROS)}.\hskip 1em plus 0.5em minus 0.4em\relax IEEE,
  2017, pp. 23--30.

\bibitem{peng2020learning}
X.~B. Peng, E.~Coumans, T.~Zhang, T.-W.~E. Lee, J.~Tan, and S.~Levine,
  ``Learning agile robotic locomotion skills by imitating animals,'' in
  \emph{Robotics: Science and Systems}, 07 2020.

\bibitem{Jungdam2020}
\BIBentryALTinterwordspacing
J.~Won, D.~Gopinath, and J.~Hodgins, ``A scalable approach to control diverse
  behaviors for physically simulated characters,'' \emph{ACM Trans. Graph.},
  vol.~39, no.~4, aug 2020. [Online]. Available:
  \url{https://doi.org/10.1145/3386569.3392381}
\BIBentrySTDinterwordspacing

\bibitem{peng2021amp}
X.~B. Peng, Z.~Ma, P.~Abbeel, S.~Levine, and A.~Kanazawa, ``Amp: Adversarial
  motion priors for stylized physics-based character control,'' \emph{ACM
  Trans. Graph.}, vol.~40, no.~4, 2021.

\bibitem{jain2019hierarchical}
D.~Jain, A.~Iscen, and K.~Caluwaerts, ``Hierarchical reinforcement learning for
  quadruped locomotion,'' in \emph{2019 IEEE/RSJ International Conference on
  Intelligent Robots and Systems (IROS)}, 2019, pp. 7551--7557.

\bibitem{yang2021fast}
Y.~Yang, T.~Zhang, E.~Coumans, J.~Tan, and B.~Boots, ``Fast and efficient
  locomotion via learned gait transitions,'' in \emph{5th Annual Conference on
  Robot Learning}, 2021.

\bibitem{starke2019nsm}
S.~Starke, H.~Zhang, T.~Komura, and J.~Saito, ``Neural state machine for
  character-scene interactions,'' \emph{ACM Trans. Graph.}, vol.~38, no.~6,
  Nov. 2019.

\bibitem{starke2020localphase}
S.~Starke, Y.~Zhao, T.~Komura, and K.~Zaman, ``Local motion phases for learning
  multi-contact character movements,'' \emph{ACM Trans. Graph.}, vol.~39,
  no.~4, Jul. 2020.

\bibitem{bergamin2019drecon}
K.~Bergamin, S.~Clavet, D.~Holden, and J.~R. Forbes, ``Drecon: Data-driven
  responsive control of physics-based characters,'' \emph{ACM Trans. Graph.},
  vol.~38, no.~6, p. 206, 2019.

\bibitem{park2019learning}
S.~Park, H.~Ryu, S.~Lee, S.~Lee, and J.~Lee, ``Learning predict-and-simulate
  policies from unorganized human motion data,'' \emph{ACM Trans. Graph.},
  vol.~38, no.~6, Nov. 2019.

\bibitem{kober2014movement}
J.~Kober, J.~Peters, J.~Kober, and J.~Peters, ``Movement templates for learning
  of hitting and batting,'' \emph{Learning Motor Skills: From Algorithms to
  Robot Experiments}, pp. 69--82, 2014.

\bibitem{pastor2009learning}
P.~Pastor, H.~Hoffmann, T.~Asfour, and S.~Schaal, ``Learning and generalization
  of motor skills by learning from demonstration,'' in \emph{2009 IEEE
  International Conference on Robotics and Automation}.\hskip 1em plus 0.5em
  minus 0.4em\relax IEEE, 2009, pp. 763--768.

\bibitem{konidaris2009skill}
G.~Konidaris and A.~Barto, ``Skill discovery in continuous reinforcement
  learning domains using skill chaining,'' \emph{Advances in neural information
  processing systems}, vol.~22, 2009.

\bibitem{lee2019learning}
Y.~Lee, J.~Yang, and J.~J. Lim, ``Learning to coordinate manipulation skills
  via skill behavior diversification,'' in \emph{International conference on
  learning representations}, 2019.

\bibitem{clegg2018learning}
A.~Clegg, W.~Yu, J.~Tan, C.~K. Liu, and G.~Turk, ``Learning to dress:
  Synthesizing human dressing motion via deep reinforcement learning,''
  \emph{ACM Transactions on Graphics (TOG)}, vol.~37, no.~6, pp. 1--10, 2018.

\bibitem{christmann2023expanding}
G.~Christmann, Y.-S. Luo, J.~H. Soeseno, and W.-C. Chen, ``Expanding
  versatility of agile locomotion through policy transitions using latent state
  representation,'' \emph{arXiv preprint arXiv:2306.08224}, 2023.

\bibitem{haynes2011gait}
G.~C. Haynes, F.~R. Cohen, and D.~E. Koditschek, ``Gait transitions for
  quasi-static hexapedal locomotion on level ground,'' in \emph{Robotics
  Research}.\hskip 1em plus 0.5em minus 0.4em\relax Springer, 2011, pp.
  105--121.

\bibitem{boussema2019online}
C.~Boussema, M.~J. Powell, G.~Bledt, A.~J. Ijspeert, P.~M. Wensing, and S.~Kim,
  ``Online gait transitions and disturbance recovery for legged robots via the
  feasible impulse set,'' \emph{IEEE Robotics and automation letters}, vol.~4,
  no.~2, pp. 1611--1618, 2019.

\bibitem{badnava2023new}
B.~Badnava, M.~Esmaeili, N.~Mozayani, and P.~Zarkesh-Ha, ``A new
  potential-based reward shaping for reinforcement learning agent,'' in
  \emph{2023 IEEE 13th Annual Computing and Communication Workshop and
  Conference (CCWC)}.\hskip 1em plus 0.5em minus 0.4em\relax IEEE, 2023, pp.
  01--06.

\bibitem{makoviychuk2021isaac}
V.~Makoviychuk, L.~Wawrzyniak, Y.~Guo, M.~Lu, K.~Storey, M.~Macklin,
  D.~Hoeller, N.~Rudin, A.~Allshire, A.~Handa \emph{et~al.}, ``Isaac gym: High
  performance gpu-based physics simulation for robot learning,'' \emph{arXiv
  preprint arXiv:2108.10470}, 2021.

\bibitem{schulman2017ppo}
J.~Schulman, F.~Wolski, P.~Dhariwal, A.~Radford, and O.~Klimov, ``Proximal
  policy optimization algorithms,'' \emph{arXiv preprint arXiv:1707.06347},
  2017.

\bibitem{Margolis-RSS-22}
G.~Margolis, G.~Yang, K.~Paigwar, T.~Chen, and P.~Agrawal, ``{Rapid Locomotion
  via Reinforcement Learning},'' in \emph{Proceedings of Robotics: Science and
  Systems}, New York City, NY, USA, June 2022.

\bibitem{kumar2021rma}
A.~Kumar, Z.~Fu, D.~Pathak, and J.~Malik, ``Rma: Rapid motor adaptation for
  legged robots,'' in \emph{Robotics: Science and Systems}, 2021.

\end{thebibliography}

\end{document}